\documentclass[conference]{IEEEtran}
\IEEEoverridecommandlockouts
\usepackage{cite}
\usepackage{amsmath,amssymb,amsfonts}
\usepackage{algorithmic}
\usepackage{graphicx}
\usepackage{textcomp}
\usepackage{xcolor}
\usepackage[table,xcdraw]{xcolor}
\def\BibTeX{{\rm B\kern-.05em{\sc i\kern-.025em b}\kern-.08em
    T\kern-.1667em\lower.7ex\hbox{E}\kern-.125emX}}
\begin{document}

\title{Using Unwrapped Full Color Space Recording to Measure the Exposedness of Vehicle Exterior Parts for External Human Machine Interfaces}

\author{
\IEEEauthorblockN{Jose Gonzalez-Belmonte}
\IEEEauthorblockA{\textit{Electrical and Computer Engineering} \\
\textit{University of Michigan-Dearborn}\\
Dearborn, United States of America \\
josegonz@umich.com}
\and
\IEEEauthorblockN{Jaerock Kwon}
\IEEEauthorblockA{\textit{Electrical and Computer Engineering} \\
\textit{University of Michigan-Dearborn}\\
Dearborn, United States of America \\
jrkwon@umich.edu}
}

\maketitle

\begin{abstract}
One of the concerns with autonomous vehicles is their ability to communicate their intent to other road users, specially pedestrians, in order to prevent accidents. External Human-Machine Interfaces (eHMIs) are the proposed solution to this issue, through the introduction of electronic devices on the exterior of a vehicle that communicate when the vehicle is planning on slowing down or yielding. This paper uses the technique of unwrapping the faces of a mesh onto a texture where every pixel is a unique color, as well as a series of animated simulations made and ran in the Unity game engine, to measure how many times is each point on a 2015 Ford F-150 King Ranch is unobstructed to a pedestrian attempting to cross the road at a four-way intersection. By cross-referencing the results with a color-coded map of the labeled parts on the exterior of the vehicle, it was concluded that while the bumper, grill, and hood were the parts of the vehicle visible to the crossing pedestrian most often, the existence of other vehicles on the same lane that might obstruct the view of these makes them insufficient. The study recommends instead a distributive approach to eHMIs by using both the windshield and frontal fenders as simultaneous placements for these devices.
\end{abstract}


\section{Introduction}
As the promised driverless future approaches, concerns about the ability of Autonomous Vehicles (AVs) to communicate with pedestrians have repeatedly arisen. Some of these concerns include questions about the role of eye-contact \cite{Onkhar}, and implicit driver communication \cite{deWinter} in safe vehicle-pedestrian interactions. External Human-Machine Interfaces (eHMIs) were created as a possible solution to this issue, devices installed on the exterior of an AV to communicate to a pedestrian when it’s safe for them to cross or not, in the process reducing the ambiguity and social void created by the removal of a human driver. 

These devices, however, are not yet standardized, and proposals for them vary drastically in method, shape, objective, and most relevant to this work, placement on the vehicle. The placement of eHMIs has been studied before in its relationship to a pedestrian’s decisions to trust yielding incoming vehicles \cite{deClercq, Palmeiro, Bazilinskyy}, but few studies have approached the ability for pedestrians to see these devices based on their position on the vehicle. 

\cite{Troel-madec} used a color-based approach to measure the visibility of the different parts of three types of vehicles (a city car, a sedan, and a van), when placed in one of three positions in a line of three cars. By capturing pictures of the vehicles and alternating the pedestrian viewing direction and height, the study concluded that the presence of other vehicles in the road make the side of the car to be the most visible area of said vehicle and ideal for eHMI placement. It also presented a prototype display that focused on the windshield rails and front doors to signal pedestrians about the vehicle's intent.

\cite{Gonzalez} used a raycasting-based approach to evaluate the exposedness of fifteen different vehicles of different sizes and types when placed in a two-way two-lane street with walkways and alleyways on both sides of the road. The vehicles could be in one of eight different positions (three on each lane and one on each alleyway, driving over the sidewalk), and share the environment with up to seven other vehicles of the same model. Through manual review of the results, the study found the windshield, front fenders, and side mirrors to have the highest exposedness across all vehicle parts, it also concluded that an ideal placement for eHMIs should be distributive, in at least two separate places, and symmetrical along the length of the vehicle.   

While valuable, these two studies feature only static vehicles in sterile scenarios that do not directly reflect a real driving environment, monocular cameras, and lack granularity in their data capture. The objective of the present work is to expand on the results of both of these works. This is done by simulating multiple scenarios of a vehicle in movement and capturing the points on its exterior that are exposed to pedestrians in the process. The results of this study are then analyzed in comparison to the vehicle parts in order to make recommendations for optimal eHMI placement.

\section{Methods}

The objective of this study is to use find what parts on the exterior of a moving vehicle are not obscured to pedestrians, when these are looking for incoming vehicles to decide if they will cross the road. In order to do this, we needed to: 
\begin{enumerate}
    \item Identify a vehicle and a corresponding 3D model for the experiment.
    \item Identify an environment for the simulation to take place in and produce a 3D model of it.
    \item Select the software necessary for the preparation and execution of the simulation, then import both the vehicle and the environment into the software.
    \item Determine and implement scenarios to be tested.
    \item Design and develop a method of capturing the data.
    \item Run the simulation.
    \item Analyze and display the results.
\end{enumerate}

\subsection{Vehicle}
According to autoinsurance.com \cite{Vidgerrman} and edmunds.com \cite{Edmunds} the Ford F-Series was the best selling line of cars for 2025 with 828,832 units sold. For this reason we decided to perform this study with a 3D model of a vehicle from that series, since it represents a trend in the road that affects pedestrian safety. We specifically selected the 2015 Ford F-150 King Ranch, as a 3D model is available on Sketchfab.com \cite{Holiday}.

\subsection{Environment}
We wanted this simulation to reflect real-life challenges and conditions, so we decided to use a 3D model based on a real location. We specifically chose MCity (https://mcity.umich.edu/), a mixed-reality testing facility located on the University of Michigan campus, and its open-source digital twin available on GitHub (https://github.com/mcity/mcity-digital-twin). We chose this location due to its availability and layout design for autonomous vehicle testing. It also presented realistic buildings and obstacles that could obscure the view of a pedestrian.

\subsection{Selecting the Software}
For this study, we made use of the Unity game engine \cite{Unity}, as it provides a flexible and well-documented platform to write custom code, as well as a built-in rendering engine that can be expanded or modified based on our needs. We set the engine camera to use nearest neighbor filtering, turned off post-processing, disabled ambient occlusion, and used the Standard Render Pipeline to prevent any color alteration. We also used Blender 3.0 \cite{Blender} to make modifications to the meshes used; and Jupyter Notebook with Python 3 to generate the Full Color Space Palette seen below.

\subsection{Scenarios}
\begin{figure}[htbp]
\centerline{\includegraphics[width=3.5in,height=3.5in,clip,keepaspectratio]{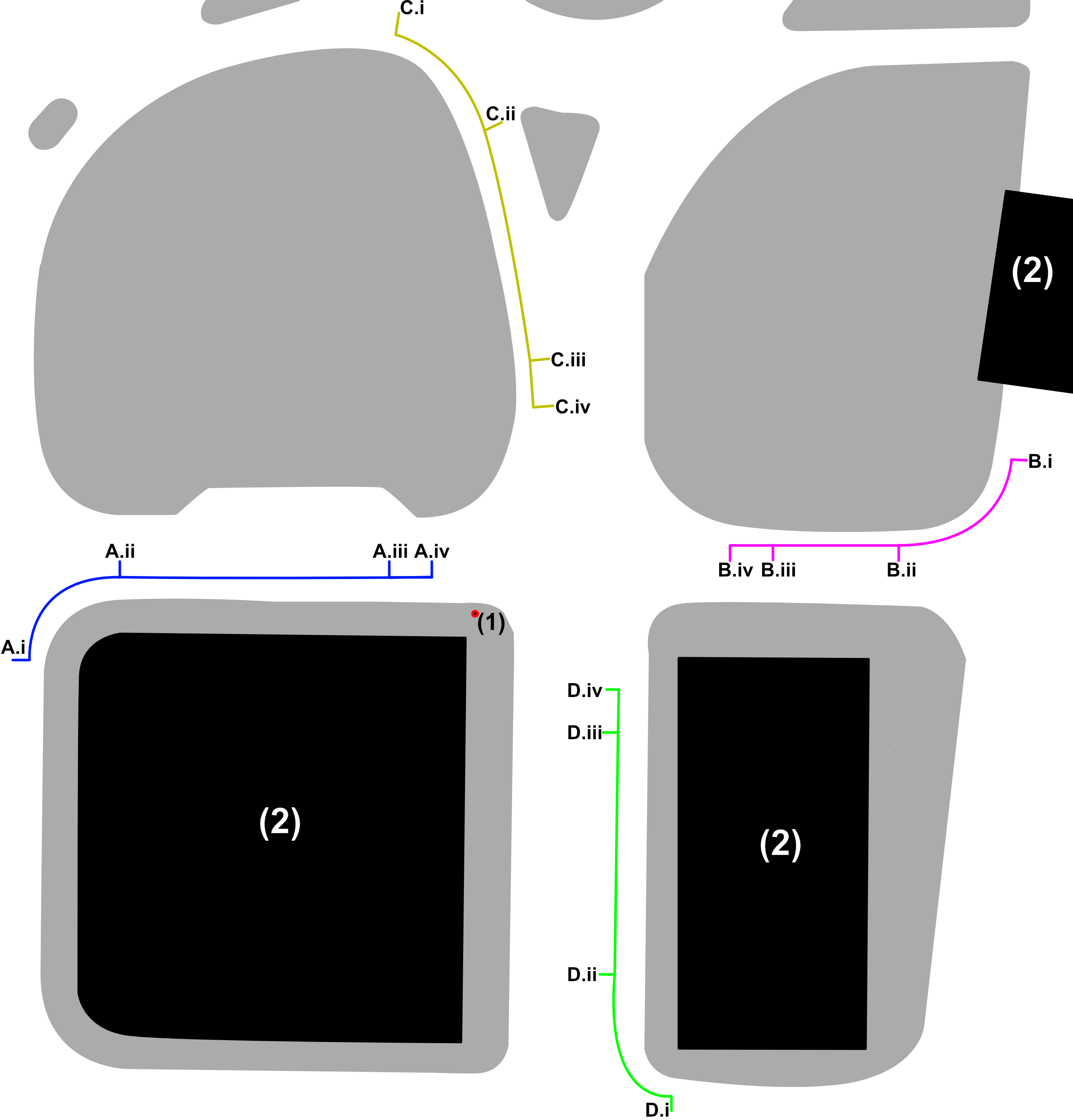}}
\caption{Illustration of the vehicle trajectory at each of the scenarios, as well as the position of the camera (1) and buildings (2). (A) Vehicle approaches from the West; (B) vehicle approaches from the East; (C) vehicle approaches from the North; (D) vehicle approaches from the South. (A-D).i is the vehicle position at the 0:00 timestamp; (A-D).ii is the vehicle position at 1:30; (A-D).iii is the vehicle position at 2:30; (A-D).iv is the vehicle position at 3:00.}
\label{Layout}
\end{figure}

For the scope of this study, we decided to focus on the visibility of vehicles that are approaching a pedestrian crossing with. Based on the layout of MCity, we identified a four-way intersection where a crossing pedestrian could face four possible directions for incoming vehicles. We placed the camera on the south-west corner of said intersection, about to cross the left crosswalk, and defined four scenarios for testing here labeled for easy reference:
\begin{itemize}
    \item Scenario A (Fig. \ref{Layout}.A.i-iv): The vehicle turns right on a corner and approaches the crossing from the West, left of the pedestrian, then stops.
    \item Scenario B (Fig. \ref{Layout}.B.i-iv): The vehicles turns right on a corner and approaches the right crossing of the intersection, then stops. The surface is slightly sloped as the vehicle approaches the intersection.
    \item Scenario C (Fig. \ref{Layout}.C.i-iv): The vehicle turns right at a roundabout, approaches the intersection from its north side (front of the pedestrian), and stops at the crosswalk.
    \item Scenario D (Fig. \ref{Layout}.D.i-iv): The vehicle turns right and approaches the intersection from the south (behind the pedestrian). 
\end{itemize}

In order to simulate an average pedestrian and binocular vision, we placed two parallel cameras 1.75 m from the ground and 0.1103594 m apart. The height was based on that of the average American adult, ages 20 and older, according to the Center of Disease Control (CDC) \cite{NCHS}. The distance between cameras distance was determined by finding the average pupillary distance for men and women, which in turn was calculated from a formula based on the age \cite{MacLachlan}, and averaging them. According to the 2020-2024 census, the median age in the United States is 39.1 \cite{Wilder}, which gave us an average pupillary distance of 107.0539 mm for the median American man and 113.6649 mm for the median American woman. 

These cameras were placed in an invisible “head” object scripted to always turn towards the moving vehicle, simulating a pedestrian turning to face the incoming vehicle. This head was placed at the southeast sidewalk corner of the intersection.

We then animated the trajectories for the observed vehicle. All four animations last exactly 3 seconds, follow the same structure, and start when the vehicle is first visible to the observer.

\begin{itemize}
    \item From 0.00 to 1.30 seconds the vehicle makes a right turn into the road that puts it facing the pedestrian (Fig. \ref{Layout} A-D.i to A-D.ii). In scenarios A and D, Rectangular buildings (Fig. \ref{Layout}.2) obscure the vehicle before turning.
    \item From 1.30 to 2.30 seconds the vehicle moves linearly along the road towards the crosswalk (Fig. \ref{Layout} A-D.ii to A-D.iii)
    \item From 2.30 to 3.00 seconds the vehicle slows down to a full stop at the intersection (Fig. \ref{Layout} A-D.iii to A-D.iv)
\end{itemize}

\subsection{Data Capturing}
One of our the previous experiments in this area made use of raycasting and physics to collect and process data \cite{Gonzalez}. However, the performance expense of raycasting limited this study to a 160x90 grid, which resulted in lower resolution for the data captured and graphed. It also meant that animations were not feasible without significantly increasing time, and that output data had to be analyzed manually.

\begin{figure}[htbp]
\centerline{\includegraphics[width=3in,height=3in,clip,keepaspectratio]{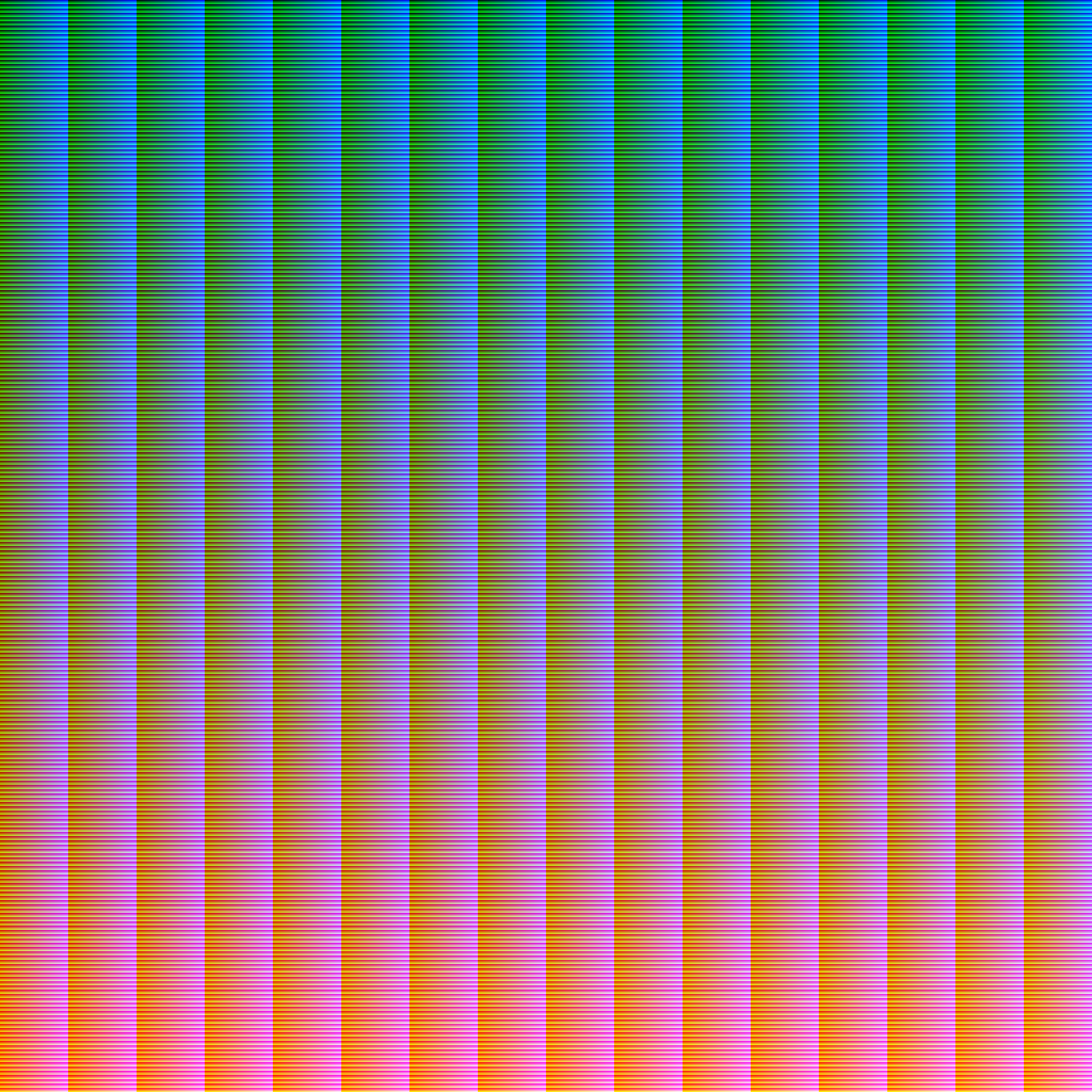}}
\caption{A rendition of Full Color Space Palette (FCSP) generated for this study, compressed for formatting purposes. Each pixel is a unique not present anywhere else in the image.}
\label{FCSP}
\end{figure}

To improve upon the previous study, we needed to implement a method to capture the visual data more efficiently and at a higher resolution without sacrificing performance. We settled on a technique that we refer to as Unwrapped Full Color Space Recording (UFCSR), which involved the following steps:

We generated a 4096x4096 image where every pixel is a unique 24-bit RGB color not present anywhere else in the image (Fig. \ref{FCSP} shows an approximation of it, compressed for formatting reasons). Since the number of pixels in the image and the number of possible color permutations in the 24-bit color space are identical ($4,096 \times 4,096 = 256\textsuperscript{3} = 16,777,216$), this means the image also includes all possible colors in that color space. While the concept of images that contain a full color space is not new, with a website \cite{Jacob} and an artbook \cite{Auberbach} dedicated to it, there does not seem to be a concise name for it, so we will refer to this image and others of its kind as Full Color Space Palettes (FCSPs).

\begin{figure}[htbp]
\centerline{\includegraphics[width=3in,height=3in,clip,keepaspectratio]{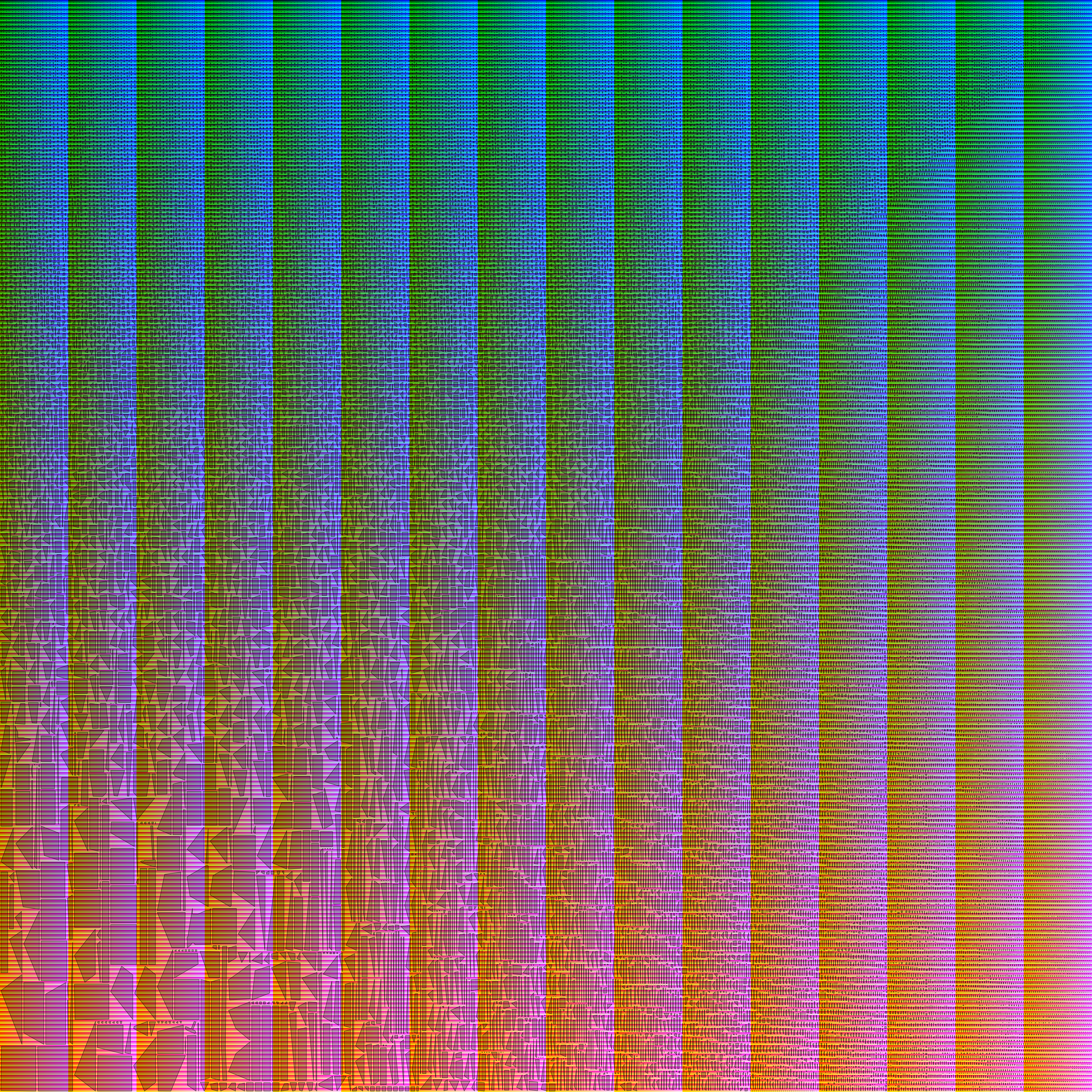}}
\caption{Faces of the F-150 King Ranch mesh unwrapped onto the FCSP}
\label{UVs}
\end{figure}

We then downloaded the 3D model of the vehicle and imported it into Blender, where we assigned a single material across all faces of the mesh. This material was set to be unlit so that its final rendered color wouldn’t be affected by shading. While still in Blender we unwrapped all faces of the mesh onto a 2D texture overlaid with the FCSP, as seen in Fig. \ref{UVs}. All edges were made to be "seams", meaning that all faces were separated and projected on the surface without maintaining contiguity. This reduced deformation and adds space between the faces to prevent sharing of colors.

\begin{figure}[htbp]
\centerline{\includegraphics[width=3in,height=3in,clip,keepaspectratio]{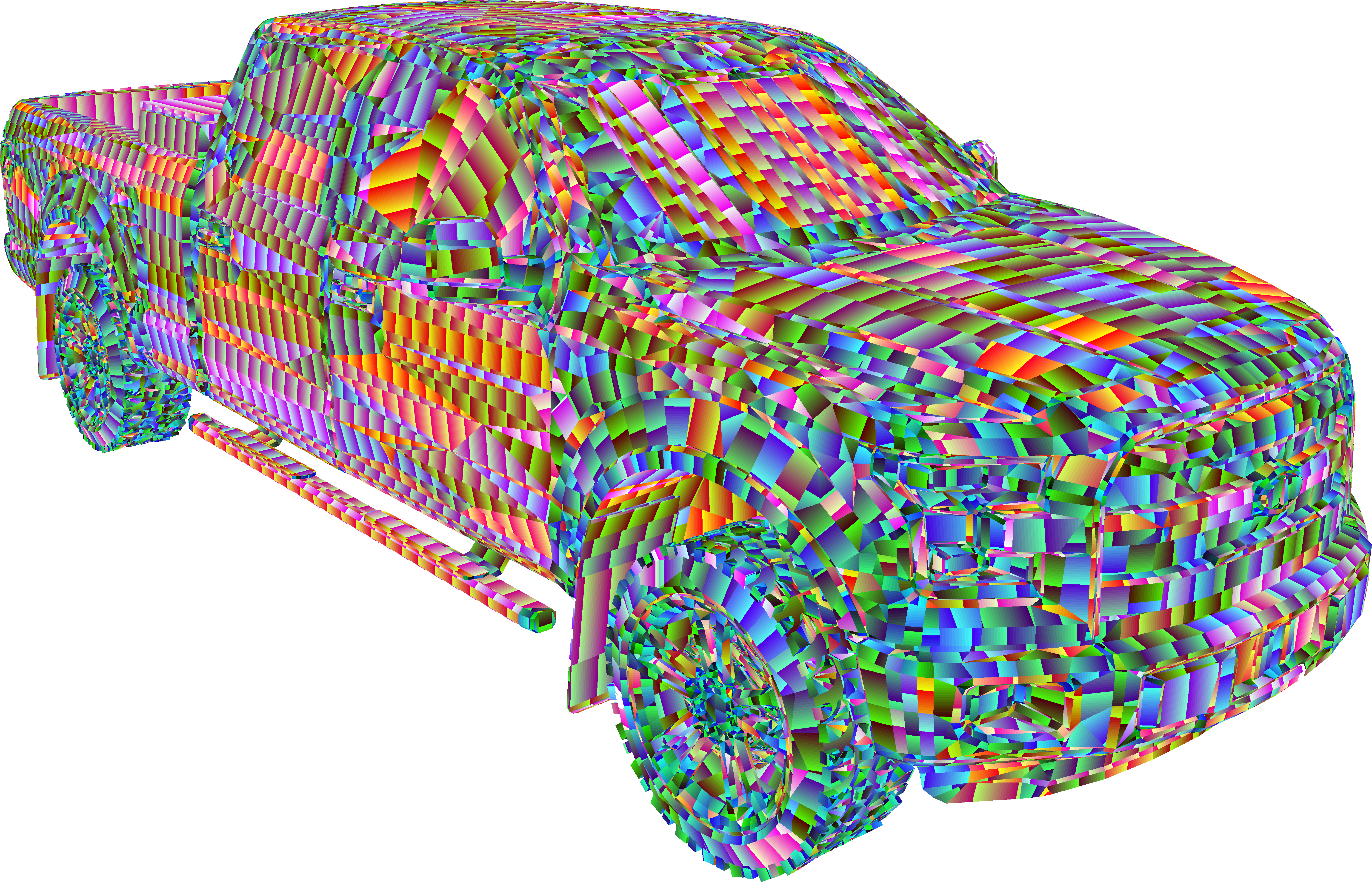}}
\caption{The vehicle model in the 3D engine with the FCSP assigned as its texture.}
\label{Textured}
\end{figure}

The 3D mesh of the vehicle was then imported into the simulation environment with its unlit material and the FCSP set as its texture (Fig. \ref{Textured})

\begin{figure}[htbp]
\centerline{\includegraphics[width=3in,height=3in,clip,keepaspectratio]{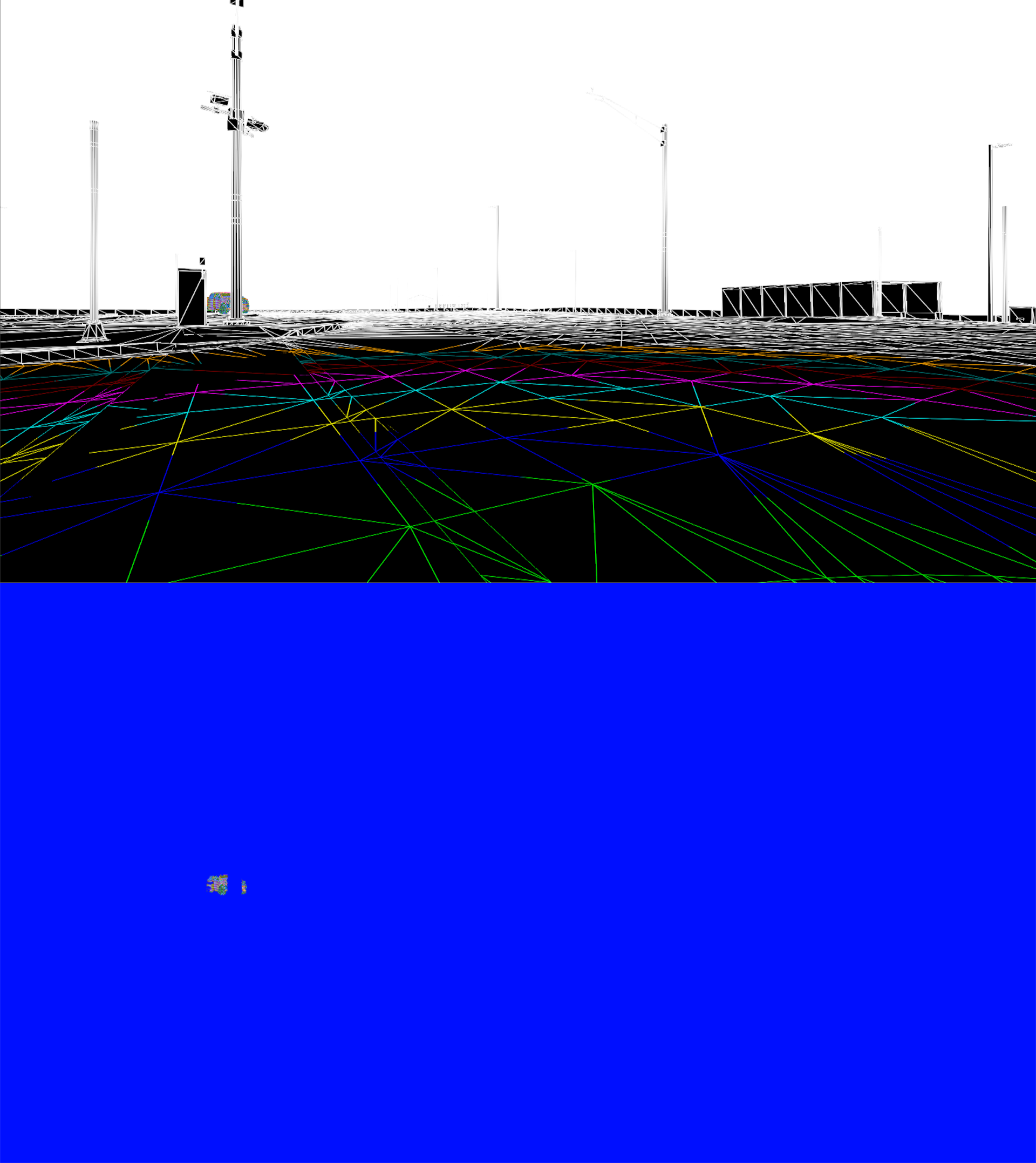}}
\caption{Comparison of the environment geometry (top) with its appearance after the unlit material has been applied to it (bottom)}
\label{Comparison}
\end{figure}

We selected a color on the FCSP not projected on any face of the vehicle (in this case, the color blue with hexadecimal code 000FFF) and assigned it as the shared unlit color for all faces of the environment that were not part of the vehicle, as well as the background. This color is set to be ignored during data processing. Fig. \ref{Comparison} shows a visual before and after this uniform color was applied.

This left us with a way to capture data on different points of the vehicle through in-engine pictures. For encoding these images we used the Portable Network Graphics (PNG) format, which is a lossless raster-graphics format\cite{Adobe}, meaning that it does not alter any of the pixel colors when compressing and hence can be analyzed pixel-by-pixel. This also allowed us to capture at a higher resolution compared to ray-casting, while significantly decreasing processing time.

In order to most closely resemble results in a real-life environment, we decided to adjust the virtual camera to match the specifications of the human eye. This involved finding its resolution, aspect ratio, and field of view. A human eye, without moving, can see 107 degrees horizontally \cite{Strasburger2020} and 135 degrees vertically \cite{Traquair}. this means the aspect ratio of an image produced by a single eye is 107:135 or approximately 3:4, with a horizontal field of view of 107 degrees. 

Regarding resolution, the human eye does not have a consistent distribution of cones and rods on its retina, making parts of its vision clearer than others. It has, however, a limited perception of between 60 and 90 pixels per degree \cite{Ashraf}, which means that the highest resolution image that a human eye can perceive ranges between $6,420\times8,100$ and $10,058\times12,690$ pixels. We decided to use the lower estimate for this study.

\begin{figure}[htbp]
\centerline{\includegraphics[width=3.5in,height=3.5in,clip,keepaspectratio]{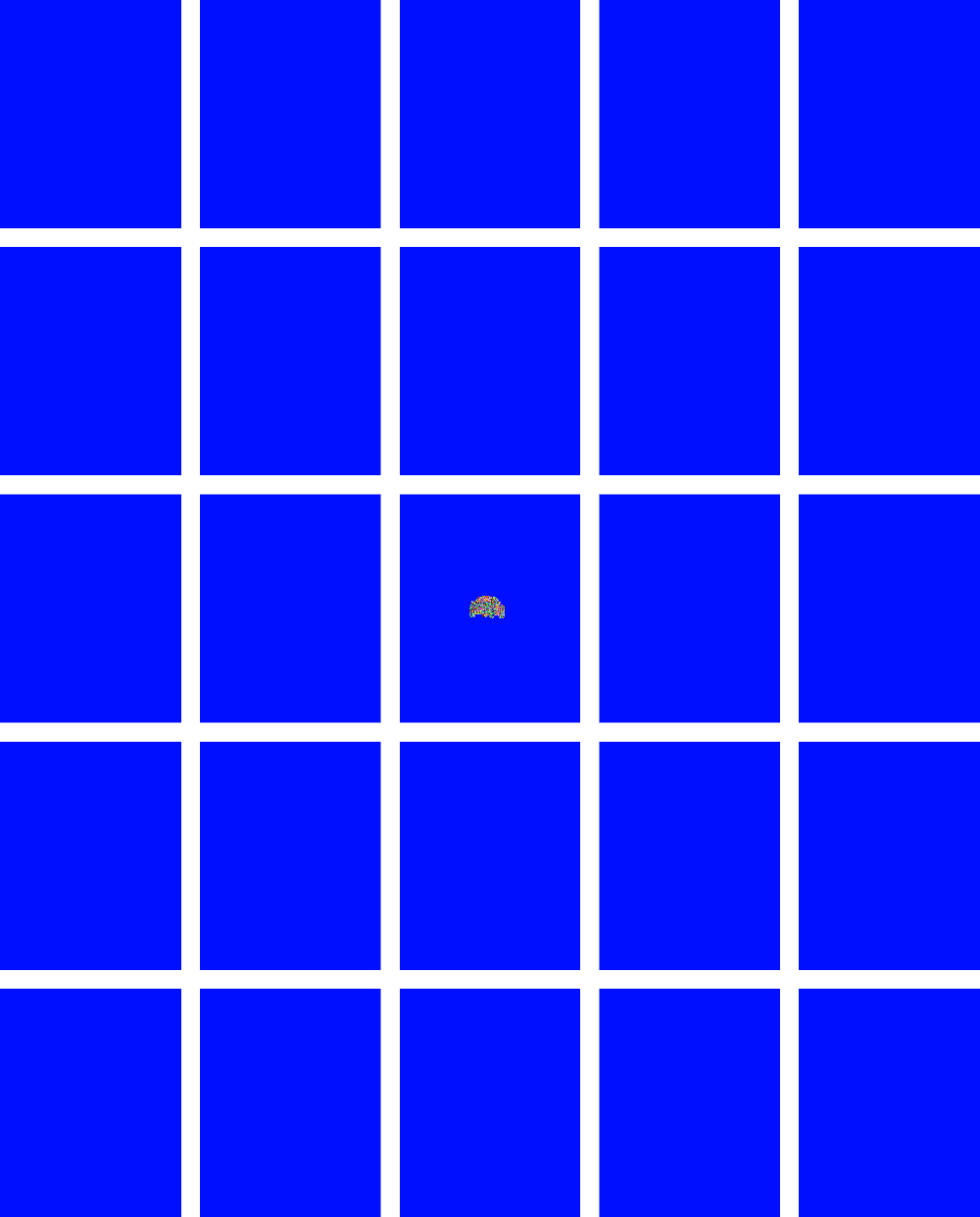}}
\caption{Compressed visualization of a $6,420\times8,100$ image being divided into twenty-five $1,284\times1,620$ files before capturing and saving.}
\label{SplitSample}
\end{figure}

To circumvent possible memory limitations we implemented a system that would split the image rendered by each camera into twenty-five equally-sized sub-segments (As seen in Fig. \ref{SplitSample}) before capturing, meaning that each capture only had to store a 1,284x1,620 file at any given time. 

During the recording of an animation, a script advances the animation for the selected scenario a specified amount of time, and has both binocular cameras alternate taking a capture of the frame at that time.

\subsection{Recording Data by Vehicle Exterior Parts}
\begin{figure}[htbp]
\centerline{\includegraphics[width=3.5in,height=3.5in,clip,keepaspectratio]{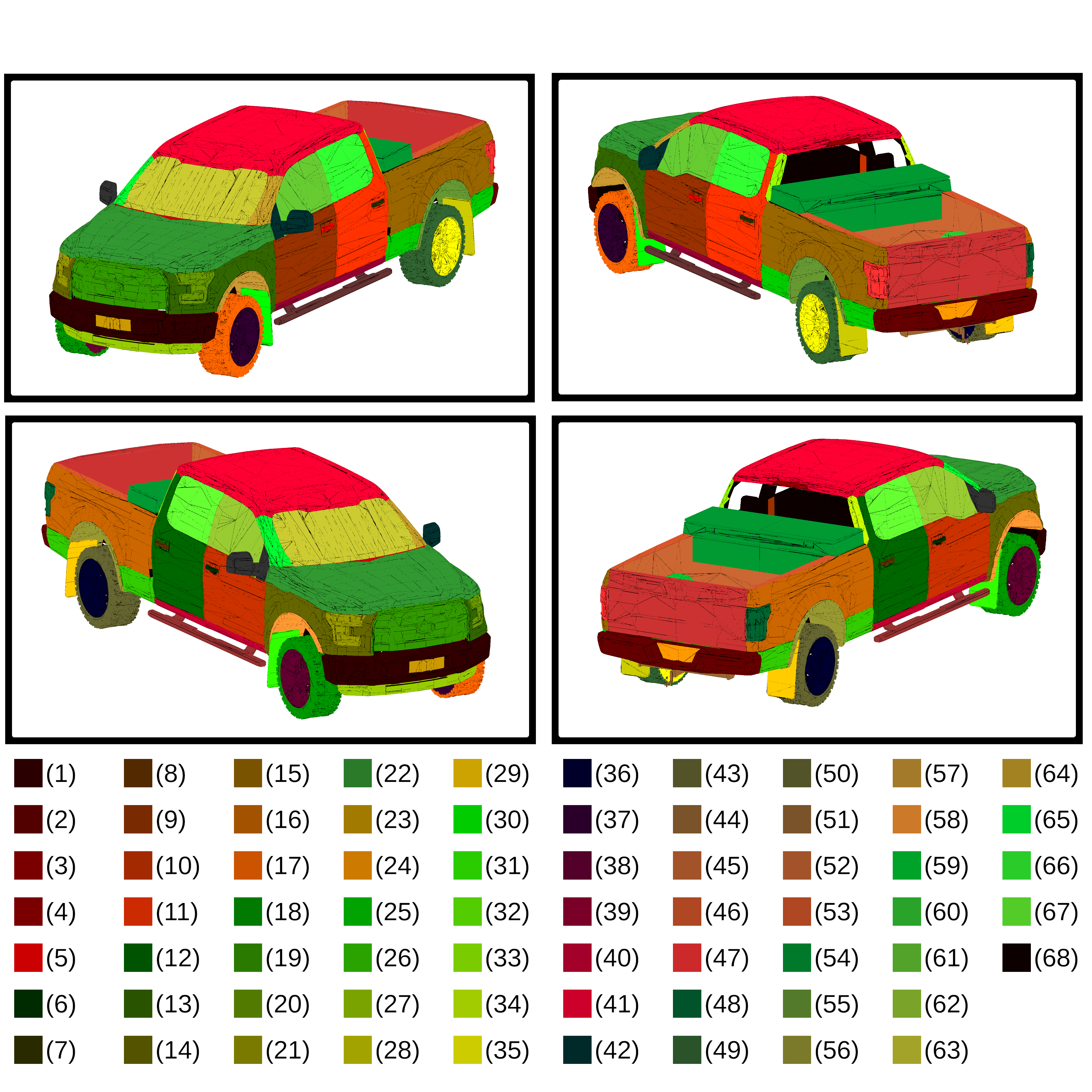}}
\caption{Exterior elements of F150 King Rach uniquely identified by color. (1) Bumper, (2) Bumper Rear, (3) Central Back Light, (4) Cowl Cover, (5) Front-Left Door Handle, (6) Front-Right Door Handle, (7) Rear-Left Door Handle, (8) Rear-Right Door Handle, (9) Front-Left Door, (10) Front-Right Door, (11) Rear-Left Door, (12) Rear-Right Door, (13) Front-Left Fender, (14) Front-Right Fender, (15) Rear-Left Fender, (16) Rear-Right Fender, (17) Front-Left Tire, (18) Front-Right Tire, (19) Grill, (20) Left Headlight, (21) Right Headlight, (22) Hood, (23) License Plate Area Front, (24) License Plate Area Rear, (25) Low Rear-Left Fender, (26) Low Rear-Right Fender, (27) Lower Reflector Front, (28) Rear-Left Mud Flap, (29) Rear-Right Mud Flap, (30) Front-Left Mud Flap, (31) Front-Right Mud Flap, (32) Rear Panel, (33) Rear-Left Panel Rail, (34) Rear-Right Panel Rail, (35) Rear-Left Rim, (36) Rear-Right Rim, (37) Front-Left Rim, (38) Front-Right Rim, (39) Left Rocker Panel, (40) Right Rocker Panel, (41) Roof, (42) Left Side Mirror, (43) Right Side Mirror, (44) Left Step Bar, (45) Right Step Bar, (46) Tailgate, (47) Left Tail Light, (48) Right Tail Light, (49) Rear-Left Tire, (50) Rear-Right Tire, (51) Trailer Hitch, (52) Truck Bed, (53) Truck Bed Rim, (54) Truck Chest, (55) Rear-Left Wheelhouse, (56) Rear-Right Wheelhouse, (57) Front-Left Wheelhouse, (58) Front-Right Wheelhouse, (59) Rear-Left Wheel Hub, (60) Rear-Right Wheel Hub, (61) Front-Left Window, (62) Front-Right Window, (63) Windshield, (64) Left Windshield Rail, (65) Right Windshield Rail, (66) Rear-Left Window, (67) Rear-Right Window, (68) Others}
\label{Parts}
\end{figure}

One of the limitations of the raycasting-based method seen in \cite{Gonzalez} is that all data had to be manually processed by a human, introducing a level of uncertainty and reducing granularity. In order to address this, we developed a method to pair UFCSR with the color-capture seen in \cite{Troel-madec}. We first mapped a different hexadecimal color to each of the different exterior elements of the vehicle, as seen in Fig. \ref{Parts}. 

Table \ref{tab:PartPixelCount} shows the amount of pixels that each vehicle exterior part takes on the texture.

\begin{table}[htbp]
\caption{Amount of pixels occupied by each vehicle exterior part on the PIdT}
\begin{center}
\begin{tabular}{|l|r|}
\hline
\textbf{Part Name} & \multicolumn{1}{l|}{\textbf{Pixel Count}} \\ \hline
Bumper & 82,655 \\ \hline
Rear Bumper & 121,050 \\ \hline
Central Back Light & 2,332 \\ \hline
Cowl Cover & 7,411 \\ \hline
Front-Left Door Handle & 2,796 \\ \hline
Front-Right Door Handle & 2,843 \\ \hline
Rear-Left Door Handle & 2,825 \\ \hline
Rear-Right Door Handle & 2,750 \\ \hline
Front-Left Door & 89,386 \\ \hline
Front-Right Door & 89,642 \\ \hline
Rear-Left Door & 49,415 \\ \hline
Rear-Right Door & 48,939 \\ \hline
Front-Left Fender & 29,422 \\ \hline
Front-Right Fender & 29,195 \\ \hline
Rear-Left Fender & 68,636 \\ \hline
Rear-Right Fender & 68,969 \\ \hline
Front-Left Tire & 199,400 \\ \hline
Front-Right Tire & 199,135 \\ \hline
Grill & 49,625 \\ \hline
Left Headlight & 18,163 \\ \hline
Right Headlight & 18,261 \\ \hline
Hood & 133,313 \\ \hline
Frontal License Plate Area & 6,849 \\ \hline
Rear License Plate Area & 3,182 \\ \hline
Low Rear-Left Fender & 24,542 \\ \hline
Low Rear-Right Fender & 32,024 \\ \hline
Frontal Lower Reflector & 19,244 \\ \hline
Rear-Left Mud Flap & 18,052 \\ \hline
Rear-Right Mud Flap & 17,683 \\ \hline
Front-Left Mud Flap & 17,899 \\ \hline
Front-Right Mud Flap & 18,429 \\ \hline
Rear Panel & 26,431 \\ \hline
Rear-Left Panel Rail & 4,148 \\ \hline
Rear-RightPanel Rail & 4,335 \\ \hline
Rear-Left Rim & 163,814 \\ \hline
Rear-Right Rim & 166,036 \\ \hline
Front-Left Rim & 157,796 \\ \hline
Front-Right Rim & 158,398 \\ \hline
Left Rocker Panel & 50,223 \\ \hline
Right Rocker Panel & 49,744 \\ \hline
Roof & 324,785 \\ \hline
Left Side Mirror & 17,758 \\ \hline
Right Side Mirror & 17,467 \\ \hline
Left Step Bar & 49,438 \\ \hline
Right Step Bar & 50,056 \\ \hline
Tailgate & 124,542 \\ \hline
Left Tail Light & 19,469 \\ \hline
Right Tail Light & 19,929 \\ \hline
Rear-Left Tire & 204,096 \\ \hline
Rear-Right Tire & 205,573 \\ \hline
Trailer Hitch & 28,855 \\ \hline
Truck Bed & 275,000 \\ \hline
Truck Bed Rim & 21,550 \\ \hline
Truck Chest & 273,960 \\ \hline
Rear-Left Wheelhouse & 32,884 \\ \hline
Rear-Right Wheelhouse & 32,855 \\ \hline
Front-Left Wheelhouse & 25,858 \\ \hline
Front-Right Wheelhouse & 26,154 \\ \hline
Rear-Left Wheel Hub & 12,942 \\ \hline
Rear-Right Wheel Hub & 12,831 \\ \hline
Front-Left Window & 22,911 \\ \hline
Front-Right Window & 22,855 \\ \hline
Windshield & 48,970 \\ \hline
Left Windshield Rail & 9,731 \\ \hline
Right Windshield Rail & 10,218 \\ \hline
Rear-Left Window & 21,811 \\ \hline
Rear-Right Window & 21,507 \\ \hline
Others & 2,849,800 \\ \hline
\end{tabular}
\label{tab:PartPixelCount}
\end{center}
\end{table}

\begin{figure}[htbp]
\centerline{\includegraphics[width=3in,height=3in,clip,keepaspectratio]{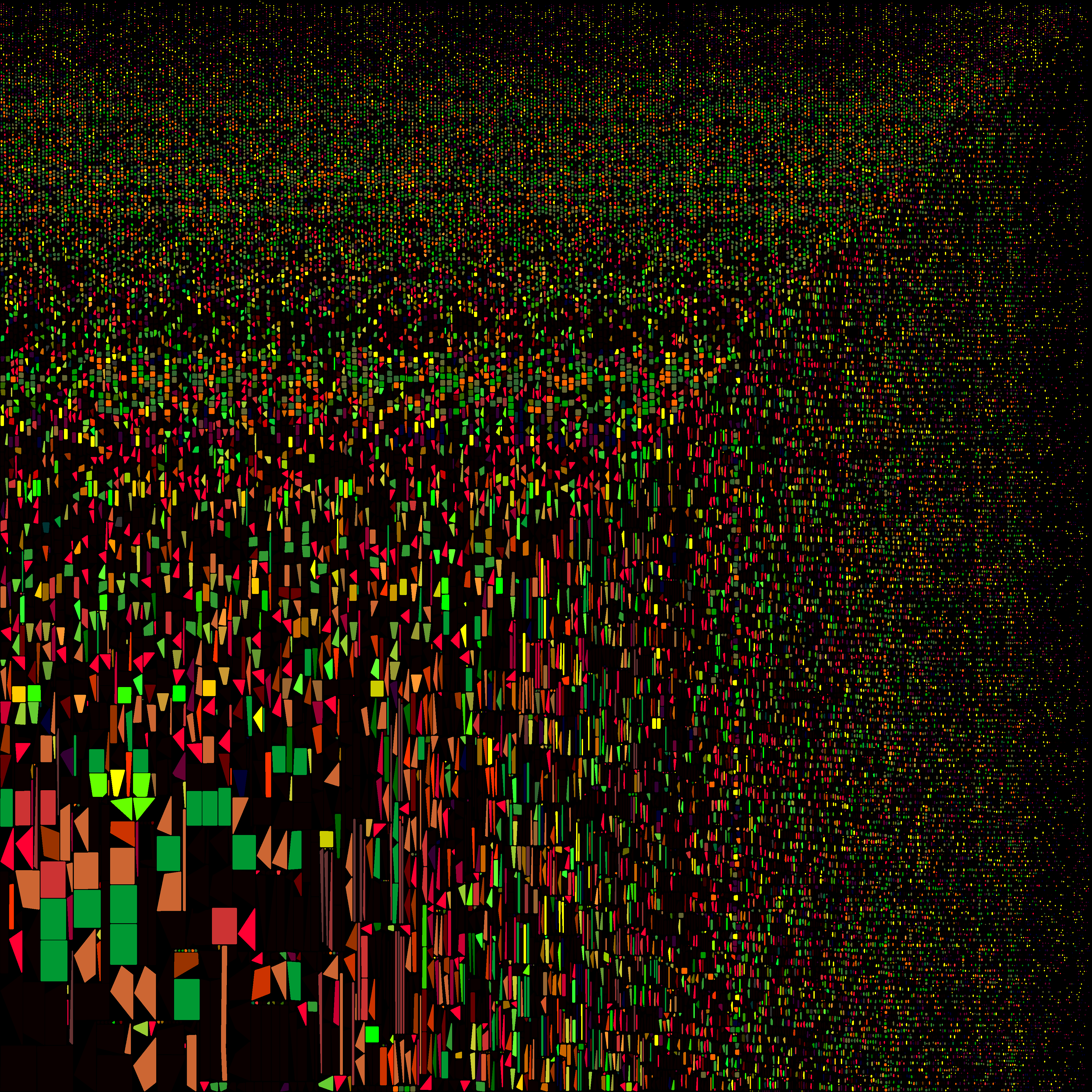}}
\caption{Unwrapped texture of the Ford F-150 King Ranch with each of its exterior elements uniquely colored. Also referred to as the Part Identification Texture (PIdT)}
\label{PartsMap}
\end{figure}

We then used Blender’s material baking functionality to paint those colors onto a “reference texture” with the same UVs and dimensions as the FCSP used in the UFCSR, meaning that any point of the FCSP could also be correlated to an exterior element of the vehicle. We refer to this texture as the Part Identification Texture (PIdT) (Fig. \ref{PartsMap}).

During processing, by comparing the values corresponding to the FCSP with the PIdT, we are able to find the number of times a part was seen, the highest value recorded for that part, the average value on it, and what percentage of the total of all values corresponds to that part.

\subsection{Trimming the Data}
Since each frame is split into smaller ones, most of the captured images are “empty”, containing only the background color which is ignored during processing. Because PNG is a lossless compressed format, images with more than one color are also of a larger file size, meaning that all “empty” images are of the same byte length (30,381 bytes for each $1,284\times1,620$ PNG file, in our case) and that any “non-empty” images would be larger than that. By ignoring all captured images under this file size threshold we improved processing time without sacrificing accuracy.

\subsection{Processing the Data}
Once the data has been captured and filtered, a different script goes through every pixel of the remaining media files and either registers its RGB value in a hash table with a value of 1, or increases the value associated with that color by 1 if it has already been registered. Hashed RGB values are only added or modified once per image to prevent any UV distortion from affecting the results. 
The hashed RGB values also correlate to that position in the texture with the exterior vehicle parts, so through this process we can also record the amount of pixels per vehicle part, the total value for all pixels of that part, the highest recorded value (which we’ll refer to as the \textit{peak value} to prevent ambiguity), and what percentage of the total of all values its made up of that part. The average value of each part is also calculated by dividing the total value of a part by the amount of pixels it occupies in the texture. All resulting data is saved in a JavaScript Object Notation (JSON) file so it can be rendered in the future without processing it again.

\subsection{Displaying the Data}
Once all files have been processed, the software generates a texture showing a heatmap of the points on the vehicle that were unobscured to the cameras the most. This is done by creating a new black $4096\times4096$ image, then going through every pixel of the original FCSP and retrieving the integer value in the hash table, in order to find how many times that color (and therefore that pixel position) was captured in that scenario. This value is then evaluated against the maximum value recorded and a plasma color gradient to find the corresponding color of the positionally corresponding pixel on the output texture. The result is a new texture with the same dimensions as the FCSP that can be applied to the mesh in order to project the evaluation of the data onto it.

\begin{figure}[htbp]
\centerline{\includegraphics[width=3in,height=3in,clip,keepaspectratio]{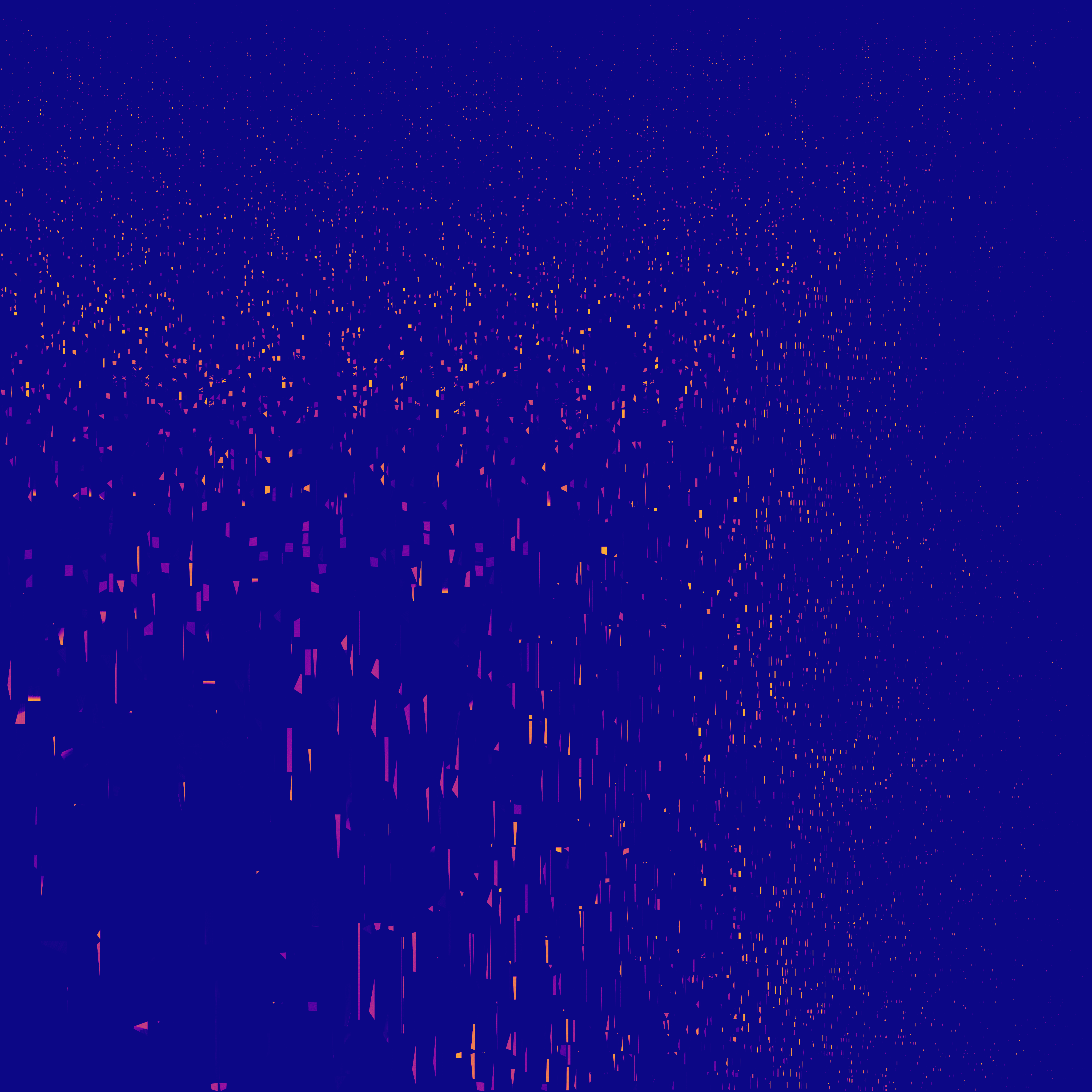}}
\caption{Example of an output plasma gradient texture.}
\label{SampleHeatmap}
\end{figure}

\subsection{Experimental Design}
The computer used for the data capturing and processing was a desktop personal computer equipped with an AMD Ryzen 7 2700 eight-core processor, an NVIDIA GeForce RTX 2060, and 16 GB of RAM running a 64-bit version of Windows 10. 

Each of the four scenarios was run and fully captured once. We decided to capture at a rate of 60 frames per second (approximately one frame every 0.018 seconds), as it matched the sample rate used during the creation of the animation, and captured one additional frame at the end of the animation. This resulted in a total of 362 frames captured per Scenario, 181 per camera, and each divided into 25 sub-images for a total of 9,050 lossless media files per scenario. Recording each scenario took an average of 12.75 minutes. Processing the data for each scenario took approximately 1 hour.

\section{Results and Discussion}

\subsection{Results}

\subsubsection{Exposedness Heatmap}
\begin{figure}[htbp]
\centerline{\includegraphics[width=3.5in,height=3.5in,clip,keepaspectratio]{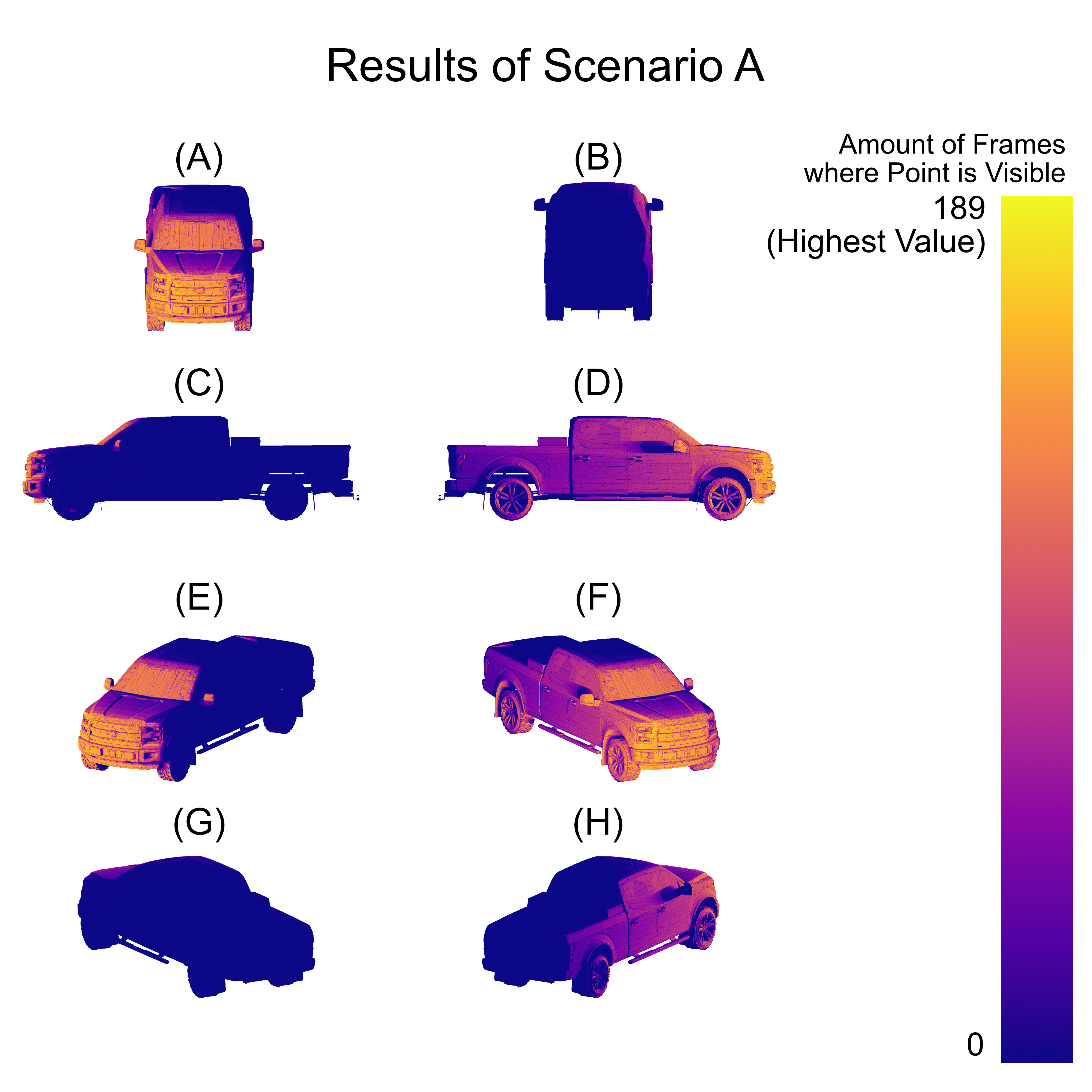}}
\caption{Pictures showing the exposedness of the exterior of the vehicle when it approaches a pedestrian from the left, stopping before a crosswalk. (A) Front, (B) Rear, (C) Left side, (D) Right Side, (E) Front-Left Side, (F) Front-Right Side, (G) Back-Left Side, (H) Back-Right Side.}
\label{AResults}
\end{figure}

\begin{figure}[htbp]
\centerline{\includegraphics[width=3.5in,height=3.5in,clip,keepaspectratio]{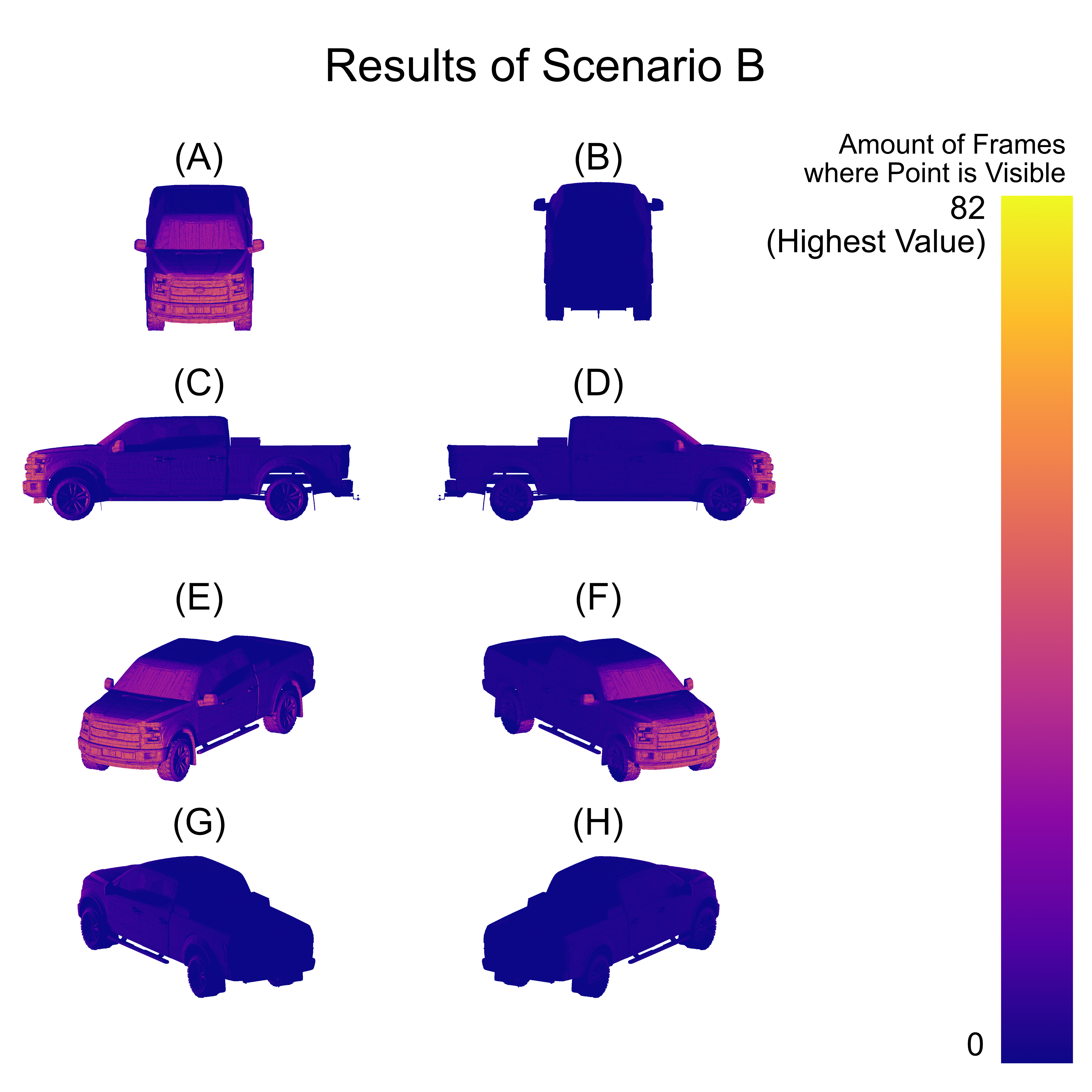}}
\caption{Pictures showing the exposedness of the exterior of the vehicle when it approaches a pedestrian from the right, stopping before a crosswalk. (A) Front, (B) Rear, (C) Left side, (D) Right Side, (E) Front-Left Side, (F) Front-Right Side, (G) Back-Left Side, (H) Back-Right Side.}
\label{BResults}
\end{figure}

\begin{figure}[htbp]
\centerline{\includegraphics[width=3.5in,height=3.5in,clip,keepaspectratio]{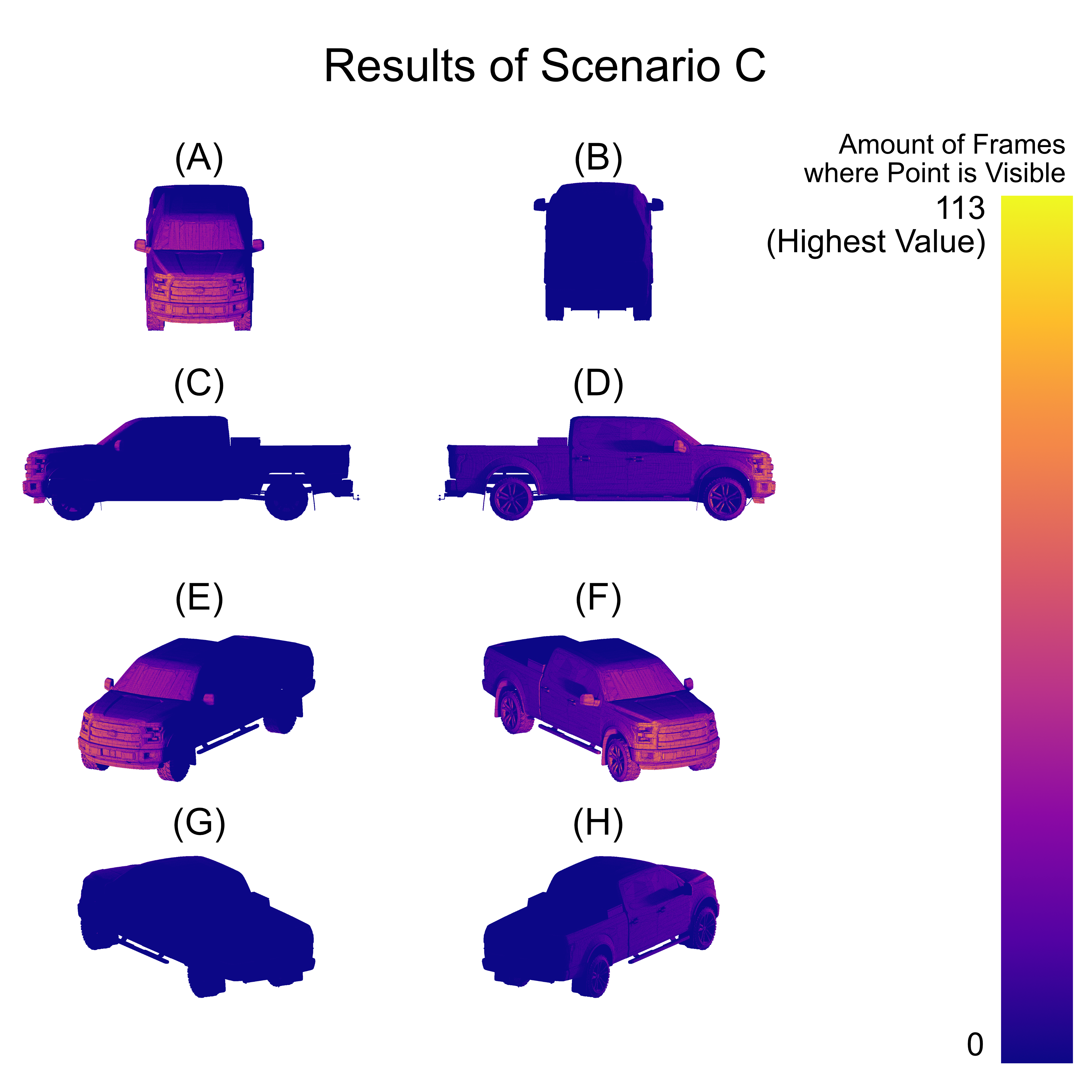}}
\caption{Pictures showing the exposedness of the exterior of the vehicle when it approaches a pedestrian from the front, stopping before a crosswalk. (A) Front, (B) Rear, (C) Left side, (D) Right Side, (E) Front-Left Side, (F) Front-Right Side, (G) Back-Left Side, (H) Back-Right Side.}
\label{CResults}
\end{figure}

\begin{figure}[htbp]
\centerline{\includegraphics[width=3.5in,height=3.5in,clip,keepaspectratio]{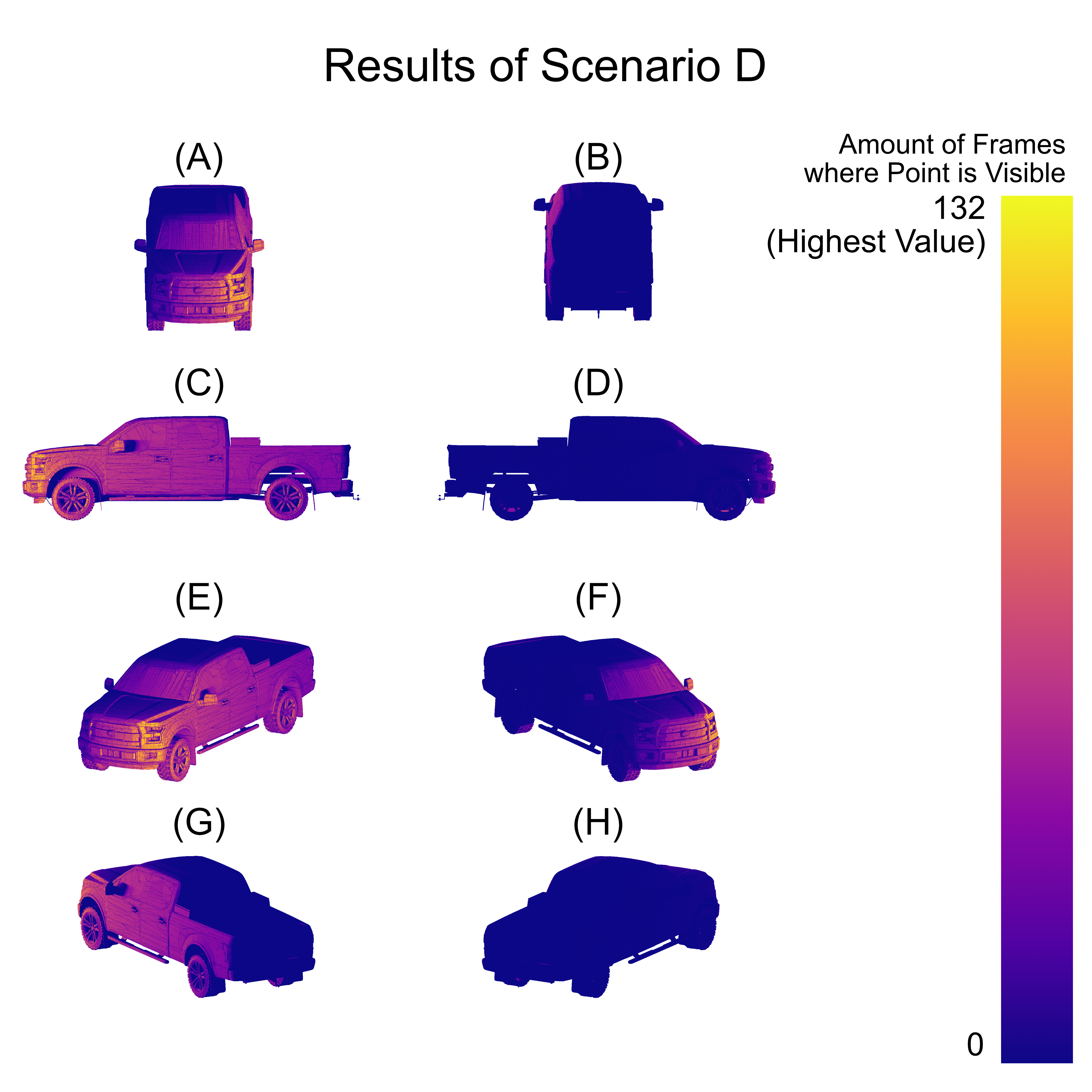}}
\caption{Pictures showing the exposedness of the exterior of the vehicle when it’s approaching a pedestrian from behind, stopping before a crosswalk. (A) Front, (B) Rear, (C) Left side, (D) Right Side, (E) Front-Left Side, (F) Front-Right Side, (G) Back-Left Side, (H) Back-Right Side.}
\label{DResults}
\end{figure}

After analyzing the images captured through Unwrapped Full Color Space Recording (UFCSR) and processing the data with the Full Color Space Palette (FCSP), we got a series of pictures (Fig. \ref{AResults}-\ref{DResults}) illustrating what parts of the vehicle exterior are visible most often across each of the four scenarios outlined in Fig. \ref{Layout}. 

\subsubsection{Vehicle Exterior Part Individual Analysis}
By processing the PIdT, we were able to produce four tables indicating which vehicle exterior parts were exposed in each scenario. To narrow down our results, for each scenario recorded we collected the five vehicle exterior parts with the highest total value (the sum of the value registered for all pixels of that part), the five with the highest average value (obtained by dividing the total by the number of pixels in that part), and the five with the highest value in any of their points (which we refer to as \textit{peak value}). The union of the three subsets is shown in Table \ref{AResults}, \ref{BResults}, \ref{CResults}, and \ref{DResults}.

\begin{table}[htbp]
\caption{Individual vehicle exterior parts that are among the five with the highest total value, average value, or peak value, for Scenario A where the vehicle is coming from the west.}
\label{tab:IndividualResultsA}
\resizebox{\columnwidth}{!}{%
\begin{tabular}{|
>{\columncolor[HTML]{EFEFEF}}l |r|r|r|r|}
\hline
\textbf{Part Name} & \multicolumn{1}{l|}{\cellcolor[HTML]{EFEFEF}\textbf{Total Value}} & \multicolumn{1}{l|}{\cellcolor[HTML]{EFEFEF}\textbf{Peak Value}} & \multicolumn{1}{l|}{\cellcolor[HTML]{EFEFEF}\textbf{Average Value}} & \multicolumn{1}{l|}{\cellcolor[HTML]{EFEFEF}\textbf{Portion of Grand Total}} \\ \hline
Bumper & \cellcolor[HTML]{EB938C}3,095,258 & \cellcolor[HTML]{68C296}180 & \cellcolor[HTML]{FFEDBB}37.45 & 9.58\% \\ \hline
Front-Right Door & \cellcolor[HTML]{F3C0BC}1,986,597 & \cellcolor[HTML]{9BD7BA}152 & \cellcolor[HTML]{FFFBEE}22.16 & 6.15\% \\ \hline
Front-Right Fender & \cellcolor[HTML]{F7D4D1}1,502,218 & \cellcolor[HTML]{66C195}181 & \cellcolor[HTML]{FFE18C}51.45 & 4.65\% \\ \hline
Front-Right Wheelhouse & \cellcolor[HTML]{FFFFFF}445,195 & \cellcolor[HTML]{68C296}180 & \cellcolor[HTML]{FFFFFF}17.02 & 1.38\% \\ \hline
Frontal Lower Reflector & \cellcolor[HTML]{FDF1F0}798,457 & \cellcolor[HTML]{77C8A0}172 & \cellcolor[HTML]{FFEAAE}41.49 & 2.47\% \\ \hline
Grill & \cellcolor[HTML]{F0ADA7}2,474,255 & \cellcolor[HTML]{71C69C}175 & \cellcolor[HTML]{FFE292}49.86 & 7.66\% \\ \hline
\cellcolor[HTML]{F3F3F3}Hood & \cellcolor[HTML]{E67C73}3,649,823 & \cellcolor[HTML]{57BB8A}189 & \cellcolor[HTML]{FFF6DD}27.38 & 11.30\% \\ \hline
Rear-Right Window & \cellcolor[HTML]{FCEDEB}903,325 & \cellcolor[HTML]{FFFFFF}97 & \cellcolor[HTML]{FFE9AC}42.00 & 2.80\% \\ \hline
Right Headlight & \cellcolor[HTML]{FFFCFC}531,741 & \cellcolor[HTML]{68C296}180 & \cellcolor[HTML]{FFF5D7}29.12 & 1.65\% \\ \hline
Windshield & \cellcolor[HTML]{EB948D}3,070,682 & \cellcolor[HTML]{94D4B4}156 & \cellcolor[HTML]{FFD666}62.71 & 9.51\% \\ \hline
\end{tabular}%
}
\end{table}
\begin{table}[htbp]
\begin{center}
\caption{Individual vehicle exterior parts that are among the five with the highest total value, average value, or peak value, for Scenario B where the vehicle is coming from the east.}
\label{tab:IndividualResultsB}
\resizebox{\columnwidth}{!}{%
\begin{tabular}{|
>{\columncolor[HTML]{EFEFEF}}l |r|r|r|r|}
\hline
\textbf{Part Name} & \multicolumn{1}{l|}{\cellcolor[HTML]{EFEFEF}\textbf{Total Value}} & \multicolumn{1}{l|}{\cellcolor[HTML]{EFEFEF}\textbf{Peak Value}} & \multicolumn{1}{l|}{\cellcolor[HTML]{EFEFEF}\textbf{Average Value}} & \multicolumn{1}{l|}{\cellcolor[HTML]{EFEFEF}\textbf{Portion of Grand Total}} \\ \hline
Bumper & \cellcolor[HTML]{E67C73}778,682 & \cellcolor[HTML]{91D3B3}70 & \cellcolor[HTML]{FFE394}9.421 & 13.46\% \\ \hline
Front-Right Fender & \cellcolor[HTML]{FDF3F2}140,031 & \cellcolor[HTML]{87CFAC}72 & \cellcolor[HTML]{FFF8E3}4.796 & 2.42\% \\ \hline
Frontal Lower Reflector & \cellcolor[HTML]{FAE5E3}216,195 & \cellcolor[HTML]{9BD7B9}68 & \cellcolor[HTML]{FFDA75}11.234 & 3.74\% \\ \hline
Grill & \cellcolor[HTML]{ED9E97}600,165 & \cellcolor[HTML]{57BB8A}82 & \cellcolor[HTML]{FFD666}12.094 & 10.37\% \\ \hline
Hood & \cellcolor[HTML]{F3BFBB}421,154 & \cellcolor[HTML]{B7E2CD}62 & \cellcolor[HTML]{FFFFFF}3.159 & 7.28\% \\ \hline
Right Headlight & \cellcolor[HTML]{FFFFFF}74,612 & \cellcolor[HTML]{9BD7B9}68 & \cellcolor[HTML]{FFFBF0}4.086 & 1.29\% \\ \hline
Windshield & \cellcolor[HTML]{EEA29C}576,115 & \cellcolor[HTML]{FFFFFF}47 & \cellcolor[HTML]{FFD86C}11.765 & 9.96\% \\ \hline
\end{tabular}%
}
\end{center}
\end{table}
\begin{table}[htbp]
\caption{Individual vehicle exterior parts that are among the five with the highest total value, average value, or peak value, for Scenario C where the vehicle is coming from the north.}
\label{tab:IndividualResultsC}
\begin{center}
\resizebox{\columnwidth}{!}{%
\begin{tabular}{|
>{\columncolor[HTML]{F3F3F3}}l |r|r|r|r|}
\hline
\cellcolor[HTML]{EFEFEF}\textbf{Part Name} & \multicolumn{1}{l|}{\cellcolor[HTML]{EFEFEF}\textbf{Total Value}} & \multicolumn{1}{l|}{\cellcolor[HTML]{EFEFEF}\textbf{Peak Value}} & \multicolumn{1}{l|}{\cellcolor[HTML]{EFEFEF}\textbf{Average Value}} & \multicolumn{1}{l|}{\cellcolor[HTML]{EFEFEF}\textbf{Portion of Grand Total}} \\ \hline
Bumper & \cellcolor[HTML]{E67C73}974,368 & \cellcolor[HTML]{AFDFC7}87 & \cellcolor[HTML]{FFE292}11.79 & 12.03\% \\ \hline
Front-Right Fender & \cellcolor[HTML]{F8D7D4}359,132 & \cellcolor[HTML]{72C69D}105 & \cellcolor[HTML]{FFE18C}12.30 & 4.43\% \\ \hline
Front-Right Rim & \cellcolor[HTML]{FEF8F7}135,267 & \cellcolor[HTML]{9ED8BC}92 & \cellcolor[HTML]{FFFFFF}0.85 & 1.67\% \\ \hline
Frontal License Plate Area & \cellcolor[HTML]{FFFFFF}82,718 & \cellcolor[HTML]{C0E6D3}82 & \cellcolor[HTML]{FFE18F}12.08 & 1.02\% \\ \hline
Frontal Lower Reflector & \cellcolor[HTML]{FAE3E1}274,805 & \cellcolor[HTML]{BCE4D1}83 & \cellcolor[HTML]{FFDB79}14.28 & 3.39\% \\ \hline
Grill & \cellcolor[HTML]{EC9A93}773,224 & \cellcolor[HTML]{A8DCC3}89 & \cellcolor[HTML]{FFD86B}15.58 & 9.54\% \\ \hline
Hood & \cellcolor[HTML]{EFA8A2}677,068 & \cellcolor[HTML]{57BB8A}113 & \cellcolor[HTML]{FFF4D5}5.08 & 8.36\% \\ \hline
Rear-Right Fender & \cellcolor[HTML]{F7D3D0}386,771 & \cellcolor[HTML]{FFFFFF}63 & \cellcolor[HTML]{FFF3D0}5.61 & 4.77\% \\ \hline
Right Headlight & \cellcolor[HTML]{FEF7F7}138,945 & \cellcolor[HTML]{A5DBC0}90 & \cellcolor[HTML]{FFEDBC}7.61 & 1.71\% \\ \hline
Windshield & \cellcolor[HTML]{EC9891}787,178 & \cellcolor[HTML]{FFFFFF}63 & \cellcolor[HTML]{FFD666}16.07 & 9.72\% \\ \hline
\end{tabular}%
}
\end{center}
\end{table}
\begin{table}[htbp]
\caption{Individual vehicle exterior parts that are among the five with the highest total value, average value, or peak value, for Scenario D where the vehicle is coming from the south.}
\label{tab:IndividualResultsD}
\begin{center}
\resizebox{\columnwidth}{!}{%
\begin{tabular}{|
>{\columncolor[HTML]{F3F3F3}}l |r|r|r|r|}
\hline
\cellcolor[HTML]{EFEFEF}\textbf{Part Name} & \multicolumn{1}{l|}{\cellcolor[HTML]{EFEFEF}\textbf{Total Value}} & \multicolumn{1}{l|}{\cellcolor[HTML]{EFEFEF}\textbf{Peak Value}} & \multicolumn{1}{l|}{\cellcolor[HTML]{EFEFEF}\textbf{Average Value}} & \multicolumn{1}{l|}{\cellcolor[HTML]{EFEFEF}\textbf{Portion of Grand Total}} \\ \hline
Bumper & \cellcolor[HTML]{ED9D96}1,107,052 & \cellcolor[HTML]{67C295}126 & \cellcolor[HTML]{FFF0C7}13.394 & 6.55\% \\ \hline
Front-Left Door & \cellcolor[HTML]{E67C73}1,396,258 & \cellcolor[HTML]{AFDFC8}98 & \cellcolor[HTML]{FFEDBB}15.621 & 8.26\% \\ \hline
Front-Left Fender & \cellcolor[HTML]{F1B4AE}905,150 & \cellcolor[HTML]{5FBF90}129 & \cellcolor[HTML]{FFD666}30.764 & 5.36\% \\ \hline
Front-Left Rim & \cellcolor[HTML]{F9DFDD}515,810 & \cellcolor[HTML]{67C295}126 & \cellcolor[HTML]{FFFFFF}3.269 & 3.05\% \\ \hline
Front-Left Window & \cellcolor[HTML]{F7D2CF}630,434 & \cellcolor[HTML]{EDF8F3}74 & \cellcolor[HTML]{FFDB79}27.517 & 3.73\% \\ \hline
Hood & \cellcolor[HTML]{EFA9A3}999,250 & \cellcolor[HTML]{62C092}128 & \cellcolor[HTML]{FFF9E8}7.496 & 5.91\% \\ \hline
Left Headlight & \cellcolor[HTML]{FFFFFF}229,394 & \cellcolor[HTML]{57BB8A}132 & \cellcolor[HTML]{FFF2CB}12.630 & 1.36\% \\ \hline
Rear-Left Door & \cellcolor[HTML]{E98C84}1,256,779 & \cellcolor[HTML]{B2E0C9}97 & \cellcolor[HTML]{FFDE84}25.433 & 7.44\% \\ \hline
Rear-Left Fender & \cellcolor[HTML]{E78178}1,355,652 & \cellcolor[HTML]{F8FCFA}70 & \cellcolor[HTML]{FFE7A4}19.751 & 8.02\% \\ \hline
Rear-Left Window & \cellcolor[HTML]{F7D5D2}606,128 & \cellcolor[HTML]{FFFFFF}67 & \cellcolor[HTML]{FFDB77}27.790 & 3.59\% \\ \hline
\end{tabular}%
}
\end{center}
\end{table}

\subsection{Discussion}

\subsubsection{Total Value Results}
In UFCSR, each time a pixel on the FCSP is observed the value associated with it increases by one. The total value of a part hence tells us how many times any pixel on said part was seen across multiple frames. Some observations regarding the total values:
\begin{itemize}
    \item The bumper is the element with the highest total value in most tables, as seen on Table \ref{tab:IndividualResultsB} and \ref{tab:IndividualResultsC}.
    \item The hood is among the top five elements with the highest total value in all four scenarios.
    \item The grill and the windshield are among the top five elements with the highest total value in Scenario A, B, and C.
    \item For Scenario D, we see lateral elements display a higher total. Table \ref{tab:IndividualResultsD} has the front-left door, rear-left fender, and rear-left door as the three element with the highest values, in that order. This is concurrent with the findings by \cite{Troel-madec}.
    \item In Scenario A, the hood is the vehicle exterior part with the highest total. Notably, this is the scenario where the vehicle gets closer to the pedestrian.
\end{itemize}

\subsubsection{Peak Value Results}
The peak value represents the single highest value found across all pixels of a part painted in the PIdT. This tells us what is the highest number of frames that any single point of that part was exposed to the camera. We can use this data to determine which vehicle exterior part has a point of its surface that pedestrians are likely to see the most. This is the criteria used to evaluate vehicle parts in \cite{Gonzalez}.  
\begin{itemize}
    \item The three vehicle parts with the highest peak value are the hood (Scenario A and C), the grill (Scenario B), and the left headlight (Scenario D).
    \item The front-right fender has the second highest peak value in Scenarios A, B, and C. The front-left fender has the second highest peak value in Scenario D. This is consistent with the results obtained in \cite{Gonzalez}.
    \item The left or the right headlight is on the top five for the four scenarios. This is concurrent with the results in \cite{Gonzalez}. 
\end{itemize}

\subsubsection{Average Value Results}
Since the vehicle exterior parts are of different sizes, the larger ones are likely to get a higher total value. By looking at the highest average value we can address this bias. This value also tells us, on average, the number of frames where each pixel on the vehicle exterior part was seen by the pedestrian.
\begin{itemize}
    \item The three vehicle parts with the highest average value are the windshield (Scenario A and C), the grill (Scenario B), and the front-left fender (Scenario D)
    \item At least one of the fenders is in the top five for Scenarios A through D. 
    \item The grill is in the top five for Scenarios A through C.    
    \item For Scenario D, we see lateral elements display a higher average. The top five are all lateral elements (front-left fender, rear-left window, front-left window, rear-left door, and rear-left fender).
\end{itemize}

\subsubsection{Recommendations}
We have identified a series of candidates for placement of external Human-Machine Interfaces (eHMIs): The bumper, the windshield, the front fenders, the grill, the rear fender, the front door, the hood, and the headlights. Notably, the side view mirrors, wheels, and windshield rails did not stand out in either of our three categories, despite these elements being concluded to be unexposed in \cite{Gonzalez} and \cite{Troel-madec}. This, as well as the higher values of frontal elements such as the bumper, grill, and hood, may be due to the lack of occluding vehicles to obscure elements on the front of the vehicle, as well as the presence of obstructions on the sidewalk that may block view of the wheels.   

Based on this data, we can make the following recommendations:
\begin{itemize}
    \item The bumper, grill, and hood are vehicle exterior parts highly exposed to pedestrians. However, as stated in \cite{Gonzalez} and \cite{Troel-madec}, they are also likely to be obscured by other vehicles on the road.
    \item The doors are exposed to pedestrians, but mainly in frames where the vehicle is moving on the side of a road perpendicular to the pedestrian's view direction. This is likely due to the American driving direction used for this study.
    \item The front fenders and the windshield were among the top five in all scenarios for two out of the three categories. This correlates to the conclusion of \cite{Gonzalez}, where one of the proposed configurations for eHMI placement is a distributive system occupying both the windshield and the front fenders. 
\end{itemize}

\section{Conclusion}

\subsection{Limitations of the Study}
The scope of this experiment was limited to a specific model of a specific series of SUVs, using a pre-existing road environment based on American right-hand driving, and only measuring the visibility of the vehicle from the perspective of a pedestrian, with no other vehicles in front of it. 

While the technique of Unwrapped Full Color Space Recording offers versatility and short preparation, identifying the elements of a vehicle and individually assigning colors to generate a PIdT requires understanding of 3D object manipulation and materials. 

Additionally, since this method depends on the analysis of colors as they are perceived by a camera and written onto an image, it is very sensitive to any modification of the visuals. For instance, Unity’s own Universal Render Pipeline comes with ambient occlusion that darkens certain parts of the mesh, so it was mandatory that this experiment used the Standard Render Pipeline. This sensitivity also means that transparent effects and evaluating the effect of lighting are not possible with the current iteration of this method, as they will affect the color registered by the camera and result in inaccuracies. 

This method does not scale well either. If one wanted to record data on multiple elements at the same time, one would need to either make each object cover only a fraction of the FCSP to prevent overlap, or increase the color space through other means. Increasing the size of an object will also reduce its resolution, as larger faces will occupy decreasingly smaller space in the UV map and hence fit less pixels of the FCSP.

\subsection{Future Work}
Since this study was done using a 2025 Ford F-150 King Ranch, all results are solely applicable to it. Improvements upon this work may involve repeating the study with more recent vehicle models, or with different classifications of size and shape. 

While there are limitations to the UFCSP technique employed in this study, modifications can be made as long as they do not disturb color output. Viable changes include adding occluding pedestrians or vehicles, fully opaque particles that simulate rain, or using dithering to simulate fog and changes in lighting. Other environments may also be tested to more widely cover the different interactions that pedestrians may have with incoming vehicles.

The source code for this study, along with the data produced by it, can be found at https://jgonzalez-uom.github.io/ufcsr/.

\subsection{Conclusion}
In this paper we used a virtual simulation to study the exposedness of the 2015 Ford F-150 King Ranch and its vehicle exterior parts as a means to determine the location on the vehicle best suited for placing external Human Machine Interfaces (eHMIs) on modern vehicles. We did this by animating the vehicle driving towards a four-way intersection and positioning a pair of cameras at the position of a pedestrian waiting to cross one of the crosswalks. The data was obtained using the Unwrapped Full Color Space Recording (UFCSR) technique, where we generated an image that contained every possible 24-bit color (known as the Full Color Space Palette (FCSP)), unwrapped the faces of a mesh on the image so that every face only contained unique colors, and recorded frames of the vehicle in motion as lossless media files. 

Through analyzing these files, and cross-referencing them with a color-coded map of the vehicle exterior parts, we determined the bumper, grill, and hood to be the parts most present in the pedestrian vision when a vehicle is approaching; however if one were to ignore purely frontal elements and account for other obscuring vehicles on the road, as previous studies have addresses, the frontal-right fenders and the windshield are concluded to be the elements on a vehicle exterior most often unobstructed and in line of sight to pedestrians during crossing situations.

This strongly concurs with the results and observations of previous similar works, and we recommend a distributive approach to eHMIs that makes use of both of these placements simultaneously, in order to maximize their visibility to pedestrians and other road users.

The UFCSR technique and the provided source code also provide a foundation for future similar studies with different vehicle models, more vehicles on the road, different crossing conditions, or changes in lighting and weather conditions simulated through dithering and solid-colored particles. The data in this experiment should nonetheless be valuable in the design of methods of communication between autonomous vehicles and other road users, specially in improving the safety and well-being of pedestrians in our roads.

\end{document}